\documentclass[10pt,twocolumn,letterpaper]{article}

\usepackage{cvpr} % uncomment this for the final submission
\usepackage{times}
\usepackage{epsfig}
\usepackage{graphicx}
\usepackage{amsmath}
\usepackage{amssymb}
\usepackage{multirow}
\usepackage{booktabs}

\usepackage[pagebackref=true,breaklinks=true,letterpaper=true,colorlinks,bookmarks=false]{hyperref}

\begin{document}

%%%%%%%%% TITLE
\title{Is It Really You? Exploring Biometric Verification Scenarios\\ in Photorealistic Talking-Head Avatar Videos}

\author{Laura Pedrouzo-Rodriguez, Pedro Delgado-DeRobles, Luis F. Gomez, \\Ruben Tolosana, Ruben Vera-Rodriguez, Aythami Morales, Julian Fierrez\\
\normalsize Biometrics and Data Pattern Analytics Lab, Universidad Autonoma de Madrid, Spain\\
% {\tt\small laura.pedrouzo@uam.es, pedro.delgadoderobles@estudiante.uam.es,}\\
% {\tt\small \{luisf.gomez, ruben.tolosana, ruben.vera, aythami.morales, julian.fierrez\}@uam.es}
}
% For a paper whose authors are all at the same institution,
% omit the following lines up until the closing ``}''.
% Additional authors and addresses can be added with ``\and'',
% just like the second author.
% To save space, use either the email address or home page, not both
% \and
% Second Author\\
% Institution2\\
% First line of institution2 address\\
% {\tt\small secondauthor@i2.org}
% }

\maketitle

\thispagestyle{empty}

%%%%%%%%% ABSTRACT
\begin{abstract}Photorealistic talking-head avatars are becoming increasingly common in virtual meetings, gaming, and social platforms. These avatars allow for more immersive communication, but they also introduce serious security risks. One emerging threat is impersonation: an attacker can steal a user's avatar, preserving his appearance and voice,making it nearly impossible to detect its fraudulent usage by sight or sound alone. In this paper, we explore the challenge of biometric verification in such avatar-mediated scenarios. Our main question is whether an individual's facial motion patterns can serve as reliable behavioral biometrics to verify their identity when the avatar's visual appearance is a facsimile of its owner. To answer this question, we introduce a new dataset of realistic avatar videos created using a state-of-the-art one-shot avatar generation model, GAGAvatar, with genuine and impostor avatar videos. We also propose a lightweight, explainable spatio-temporal Graph Convolutional Network architecture with temporal attention pooling, that uses only facial landmarks to model dynamic facial gestures. Experimental results demonstrate that facial motion cues enable meaningful identity verification with AUC values approaching 80\%. The proposed benchmark and biometric system are available\footnote{\url{https://github.com/BiDAlab/GAGAvatar-Benchmark}} for the research community in order to bring attention to the urgent need for more advanced behavioral biometric defenses in avatar-based communication systems.
\end{abstract}

%%%%%%%%% BODY TEXT
\section{Introduction}

In recent years, there has been a surge in novel methods for generating photorealistic avatars \cite{chu2024generalizable, li2025rgbavatar, liu2025avatarartist, xu2024gaussian} that can create and animate high-quality 3D human avatars with a single image or text prompt. In parallel, industry investment in avatar technology has exploded, with companies such as Synthesia\footnote{\url{https://cnb.cx/3WfCHFt}} securing hundreds of millions of dollars in funding to scale avatar services, which are being used by more than one million users. Another notable example is Meta\footnote{\url{https://www.uploadvr.com/meta-project-warhol-codec-avatars-training-paying/}} that pays users to capture their facial expressions, speech, and gestures to train its Codec Avatars model \cite{martinez2024codec}.

\begin{figure}[t]
    \centering
    \includegraphics[width=\columnwidth]{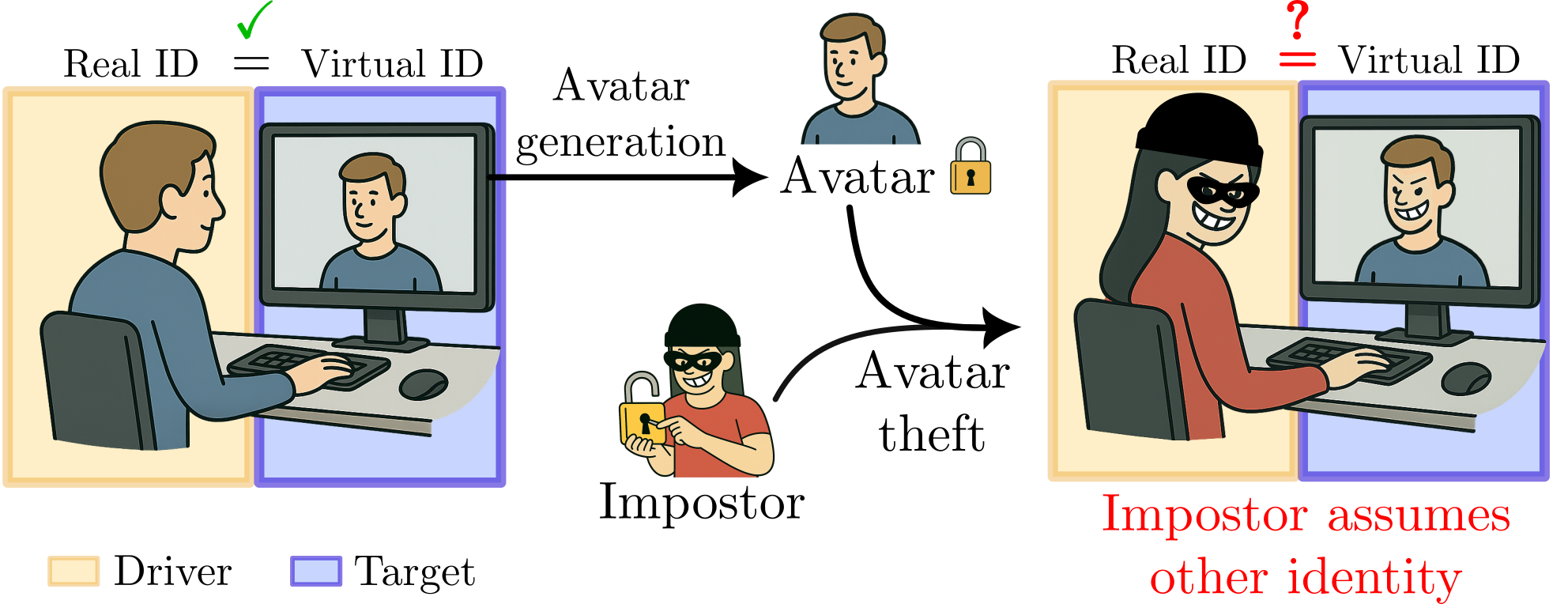}
    \caption{\textbf{Avatar impostor attack scenario:} An impostor steals a \textit{target} identity's avatar and pretends to be that identity in a virtual meeting. The visual appearance and voice of the impostor's avatar is exactly the same as the target's. The behavioral biometrics, such as facial gestures, however, are \textit{driven} by the impostor.}
    \label{fig:intro-diagram}
\end{figure}

These photorealistic talking-head avatars are becoming popular in virtual meetings, gaming, and metaverse platforms. A good example of this popularity was the virtual reality interview\footnote{\url{https://www.youtube.com/watch?v=MVYrJJNdrEg}} between Mark Zuckerberg and Lex Fridman in his podcast, which had nearly 3 million views. Although these avatars allow for immersive communication, they also pose critical security risks to society, as depicted in Fig.~\ref{fig:intro-diagram}. An impostor can steal the avatar of an identity and impersonate him or her in real time \cite{tariq2023deepfake}.

A recent example of such identity theft was a video scam in early 2024\footnote{\url{https://edition.cnn.com/2024/02/04/asia/deepfake-cfo-scam-hong-kong-intl-hnk}}, in which criminals used an AI-generated video of a company executive to fool another employee into transferring $\approx$\$25M USD to fraudulent accounts. This incident demonstrates how difficult it is to distinguish a real person from an impersonated one. Indeed, as avatar technology improves, such impostor attacks may become nearly impossible to detect by sight or sound alone. \textit{This is the specific use case considered in this study, where the avatar stolen by the impostor is so realistic in terms of appearance and voice that only behavioral biometric information, such as facial gestures, can be used to detect whether the driver identity that moves the avatar is the real owner of the avatar or an impostor}.

It is important to note that the above scenario is different from DeepFake detection \cite{tolosana2022fake-trends,tolosana2022deepfakes,verdoliva2020media}, in which the main task consists in simply predicting if a video is real or fake, not whether the person driving the avatar is, in fact, the owner of the avatar. The problem is also different (but related) to presentation attack detection (PAD) in biometrics \cite{hadid15pad,jh23pad}. Considering the differences, we can exploit methods and consolidated knowledge from both fields (DeepFakes and biometric PAD).
%\cite{rathgeb2022handbook,tolosana2020deepfakes}     % esto meterlo para el camera ready

In our study, we seek to explore biometric verification in avatar scenarios by addressing the following question: 

\begin{quote}
\textit{
     Can facial motion patterns serve as a reliable biometric characteristic to verify identity during avatar-mediated communication?
     }
\end{quote}

In the context of avatars, each person’s manner of speaking and emoting is distinctive and can be difficult to imitate. By capturing these subtle dynamics, it might be possible to verify if a photorealistic talking-head avatar is being driven by its owner or an impostor. Capturing these subtle dynamics is based on our previous work on behavioral face analysis for e-health \cite{luis21cvprw,luis23plos} and e-learning \cite{alvaro24behavior,daza25improve}.

To investigate this question, we designed an end-to-end experimental framework that simulates real-world avatar-mediated impersonation attacks while isolating facial motion as the distinguishing biometric signal. To our knowledge, the only study that has previously analyzed this scenario is \cite{prashnani2024avatar}, from NVIDIA. In their study, the authors coined the term ``\textit{avatar fingerprinting}" to refer to the detection of unauthorized use of avatars. Although their study reported promising results, we encountered the following problems when trying to reproduce it: \textit{i)} the proposed method and its corresponding code are not publicly available, limiting its reproducibility, and \textit{ii)} most research groups do not have access to the GPU resources required to train NVIDIA's verification model. Our study aims to bridge these gaps by publicly releasing our proposed avatar dataset and standard benchmark, as well as exploring lighter biometric verification approaches.

The main contributions of our research are as follows:

%Our approach consists of generating realistic avatar videos with controlled \textit{target} and \textit{driver} identities, developing a motion-based verification model that uses only facial landmarks, and evaluating the model's ability to distinguish genuine users from impostors. 

\begin{itemize}
    \item A \textbf{new public avatar video dataset} generated using a state-of-the-art one-shot avatar generator, GAGAvatar \cite{chu2024generalizable}, including genuine and impostor avatar videos. 
%To the best of our knowledge, this is the first publicly available resource that includes both genuine and impostor avatar videos specifically designed to study biometric verification in avatar scenarios.
    \item A \textbf{standard public benchmark protocol for avatar verification}, defining clear train-validation-test splits and evaluation protocols that are reproducible by the research community.
    \item A \textbf{first exploration of this challenging avatar scenario proposing a lightweight and explainable behavioral biometric system} that is based solely on the facial motion patterns of the identities. Our approach is based on a Graph Convolutional Network (GCN). Unlike \cite{prashnani2024avatar} which relies on high-dimensional pairwise distance tensors (\ie, all-pair landmark distances), our approach is well suited for facial motion modeling as it explicitly encodes the mesh-like geometry of the face through graph structure. This results in both spatial awareness (by preserving local neighborhood relationships among facial regions) and parameter efficiency, given the small and fixed graph size typical of landmark-based representations, allowing faster training and inference.

    %Our results, which approach 80\% Area Under the Curve (AUC) in the best cases, highlight the difficulty of the task.
\end{itemize}

\begin{figure*}[t]
    \centering
    \includegraphics[width=0.9\textwidth]{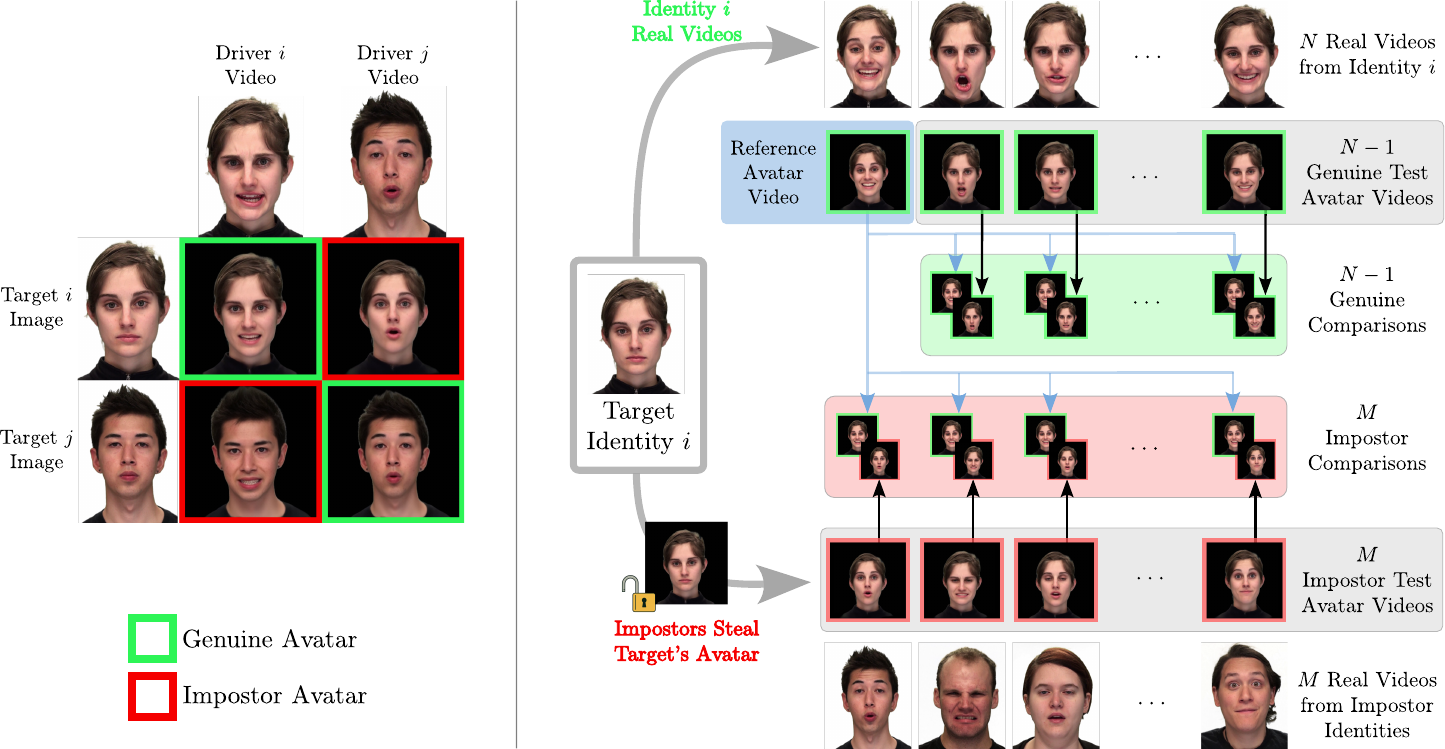}
    \caption{\textbf{Avatar generation and the proposed evaluation protocol.}  \textit{(Left)}: An avatar video is generated using a \textit{target} image (leftmost column) as the appearance for the identity, plus a \textit{driving} video (top row) to generate the facial motion in the avatar. \textit{(Right)}: For the evaluation of our proposed biometric system in avatar scenarios, we generate all \textit{genuine} and \textit{impostor comparisons} for each \textit{target} identity $i$, using all videos from $i$ and some videos from impostor identities, respectively.}
    \label{fig:avatares-pares}
\end{figure*}

The remainder of the paper is organized as follows. Sec.~\ref{sec:terminology} describes the key terminology used throughout the paper. Sec.~\ref{sec:datasets} describes the real datasets and avatar video generation. Sec.~\ref{sec:biometric_system} describes the biometric verification system proposed in the study. The experimental protocol and results are described in Sec.~\ref{sec:experimental_protocol} and Sec.~\ref{sec:experimental_result}, respectively. Sec.~\ref{sec:discussion} continues with a discussion and Sec.~\ref{sec:conclusions} concludes with key takeaways.

\section{Terminology}\label{sec:terminology}

To provide a solid understanding of our research, we first define the terminology used throughout this paper (see Fig.~\ref{fig:avatares-pares} for a visual explanation of the terminology):

\begin{itemize}
    \item \textit{Target Identity}: The person whose appearance the avatar has.
    \item \textit{Driver Identity}: The person whose facial motion controls the avatar.
    \item \textit{Avatar Video}: A video with the target's appearance and the driver's motion, generated from a \textit{target} image and a \textit{driver} video.
    \item \textit{Genuine Avatar}: When the \textit{driver} identity and the \textit{target} identity correspond to the same person, also known as \textit{self-reenactment}. Genuine avatars represent how the avatar of a real user would appear in a virtual meeting.
    \item \textit{Impostor Avatar}: When the \textit{driver} identity does not match the \textit{target} identity (also known as \textit{cross-reenactment}), \ie, when the person driving the movements of the avatar is not the same as the person corresponding to the avatar's appearance. These avatars simulate impersonation attacks, where an impostor uses someone else's avatar. 
    \item \textit{Genuine Comparison:} Comparing a reference video from a \textit{genuine avatar} against another test video generated using the same \textit{genuine avatar}. A biometric verification system should consider the latter as \textit{authorized}, indicating that both videos correspond to the same real identity.
    \item \textit{Impostor Comparison:} Comparing a reference video from a \textit{genuine avatar} against another test video generated using an \textit{impostor avatar} that has the same appearance (\textit{target identity}) as the reference \textit{genuine avatar} video. A biometric verification system should consider the latter as \textit{unauthorized}, indicating that its identity is not the same as the reference's.
\end{itemize}

\section{Datasets}\label{sec:datasets}
This section describes the datasets used in this study, including the generation of a dedicated avatar video dataset to capture identity-specific motion patterns.

%In this section, we introduce our proposed framework for biometric identity verification using \textit{exclusively} facial motion cues in avatar videos. We first describe the datasets employed in this study, including the generation of a dedicated avatar video dataset to capture identity-specific motion patterns. We then detail the architecture of the proposed verification system.

\begin{figure*}[t]
    \centering
    \includegraphics[width=\linewidth]{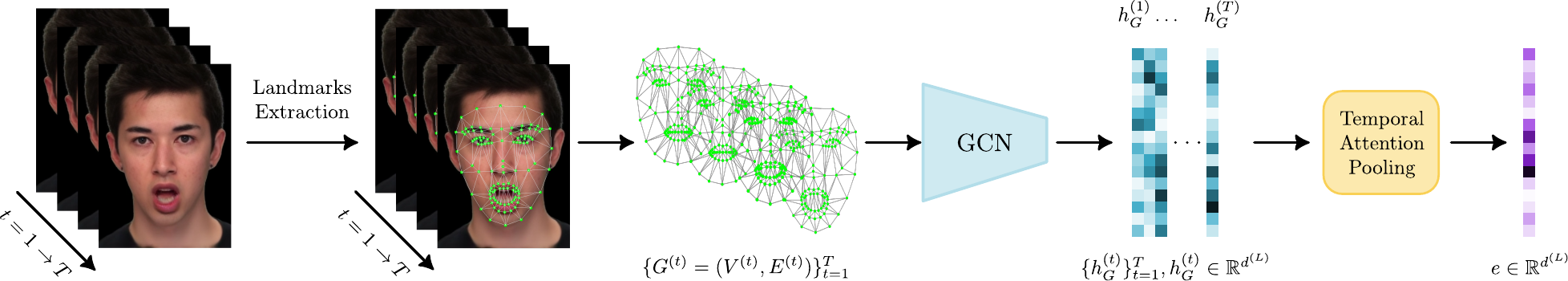}
    \caption{\textbf{Proposed biometric verification system:} For each frame $t$ in a video, a graph $G^{(t)}$ is built. All the $T$ graphs from the video are passed to the 3-layer GCN, obtaining one graph embedding $h_G^{(t)}$ per frame $t$. Finally, those $T$ graph embeddings are passed to the attention pooling module, resulting in a single embedding $e$ per video, with $e\in \mathbb{R}^{d^{(L)}}$,  $d^{(L)}$ being the output dimension of the last GCN layer, $L$.}
    \label{fig:model}
\end{figure*}

\subsection{Real datasets}

We use two public video datasets for avatar video generation. These datasets are selected because they simulate a similar environment to virtual meetings. 

\begin{itemize}
    \item \textbf{CREMA-D}\footnote{\url{https://github.com/CheyneyComputerScience/CREMA-D}} \cite{cao2014crema}: This dataset provides 7,442 short videos of 91 actors (48 male, 43 female, ages 20–74, diverse ethnicity) performing scripted sentences in various emotional states. Each actor utters 12 different sentences with one of six basic emotions (anger, disgust, fear, happy, neutral, and sad) at multiple intensity levels \cite{pena21icip}. These videos were recorded in a controlled setting with frontal views of the speaker.
    \item \textbf{RAVDESS}\footnote{\url{https://zenodo.org/records/1188976}} \cite{livingstone2018ryerson}: This dataset contains 24 actors (12 female, 12 male) speaking two fixed statements with eight emotions (calm, happy, sad, angry, fearful, surprise, disgust, and neutral). All videos were recorded in a studio environment with actors facing up against a green screen background. In total, we used 1,440 video-only speech files from the original dataset.
\end{itemize}

% Both CREMA-D and RAVDESS offer scripted, varying-intensity emotional expressions which provide a wide range of facial movements. 

% We will not use the audio files from those videos, only RGB frames.

\subsection{Avatar Video Dataset Generation}

To generate our avatar dataset, we employ the state-of-the-art avatar generation model GAGAvatar\footnote{\url{https://github.com/xg-chu/GAGAvatar}}  \cite{chu2024generalizable}. This model takes a reference image (representing the \textit{target} identity's appearance) and a driving video (providing the \textit{driver} identity's motion), and generates a new avatar video that preserves the visual appearance of the reference image while reproducing the facial expressions and movements from the driving video. The resolution of the videos generated is 512$\times$512 on a black background. It is worth noting that the GAGAvatar model does not use the audio from the original videos to generate the avatars, only the visual data. 

% \textcolor{red}{Lo digo porque los que cogen el audio seguro que siguen los movimientos de la boca mejor y saldran mas realistas}

\begin{figure}[t]
  \centering
  \begin{minipage}[b]{0.3\columnwidth}
    \centering
    \includegraphics[width=\linewidth]{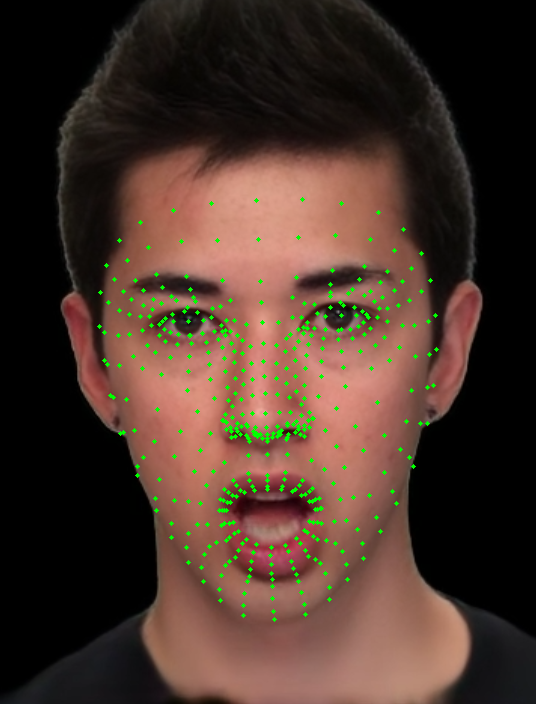}
    \vspace{1ex}
    \textbf{(a)}
  \end{minipage}
  \begin{minipage}[b]{0.3\columnwidth}
    \centering
    \includegraphics[width=\linewidth]{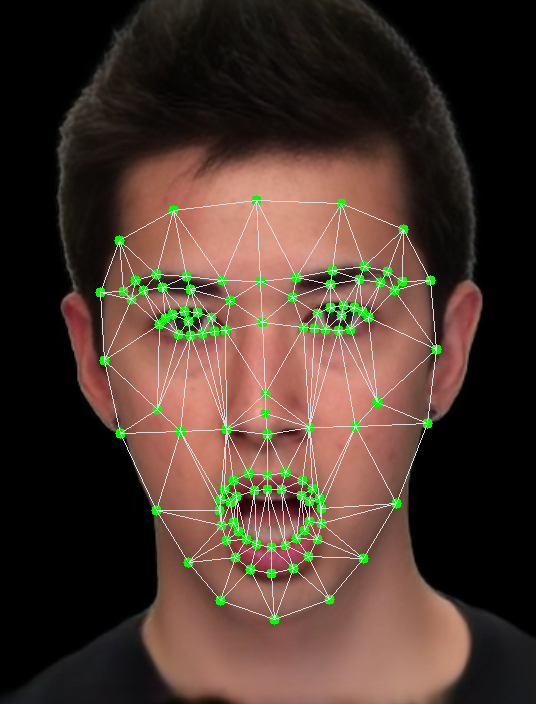}
    \vspace{1ex}
    \textbf{(b)}
  \end{minipage}
  \begin{minipage}[b]{0.3\columnwidth}
    \centering
    \includegraphics[width=\linewidth]{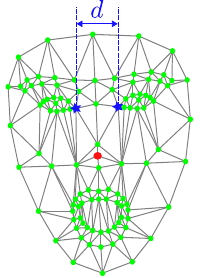}
    \vspace{1ex}
    \textbf{(c)}
  \end{minipage}
  \caption{\textbf{Landmarks extracted for each video frame:} \textit{(a)} shows all MediaPipe \cite{kartynnik2019realtimefacialsurfacegeometry} landmarks. \textit{(b)} shows the 109 selected landmarks connected via Delaunay triangulation \cite{delaunay1934sphere}. \textit{(c)} The normalization of landmarks is done subtracting the \textcolor{red}{nose tip} landmark position and scaling with \textcolor{blue}{intercanthal distance $d$} over the frames.}
  \label{fig:facemesh}
\end{figure}

For each identity in the CREMA-D and RAVDESS datasets, we first select a single, high-quality target image to represent that identity’s reference appearance. Whenever possible, we select a frontal face with neutral expression, such as in Fig.~\ref{fig:avatares-pares}, \textit{``Target Images"}. Next, we generate \textit{genuine avatar} videos using said identity's target image and all of his or her real videos as driving videos. This approach allows us to obtain new avatar videos in which the avatar's appearance corresponds to the target identity and the avatar's facial movements are genuinely made by him or her. This scenario would correspond to the typical use of the avatar by its owner in a virtual meeting.

We then generate \textit{impostor avatar} videos. For each target identity, we randomly sample a subset of videos from other identities and use them as drivers. That way, we can obtain a set of impostor avatar videos in which the appearance is the target's, while the facial movements correspond to other people, as seen in Fig.~\ref{fig:avatares-pares}.

Despite GAGAvatar being state-of-the-art, in some cases it cannot generate avatar videos that perfectly match the expressions from the driving video. For example, in Fig.~\ref{fig:avatares-pares} (left), the column corresponding to \textit{``Driver $i$"} shows that, for the avatar generated with the appearance of \textit{ ``Target $j$"}, the facial expression is subtly different from the original video. The avatar is not frowning as much and its mouth does not follow the original expression with precision.

\section{Biometric Verification System for Avatars}\label{sec:biometric_system}

Our proposed biometric system is described in Fig.~\ref{fig:model}. It is based on a Graph Convolutional Network (GCN) model, which is fed a simplified face mesh obtained from facial landmarks that are later aggregated through time via a temporal attention mechanism. Our hypothesis is that this method will be able to capture discriminative intra-frame spatial patterns of facial movement (\eg, which regions move together, muscle activation patterns), while considering their temporal variations via the attention mechanism. Next, we describe the key details of each part of the system.

\subsection{Landmarks Extraction}
\label{ssec:landmarks}

To focus on the facial motion patterns of the identities, we first extract the facial landmarks for each video using MediaPipe\footnote{\url{https://ai.google.dev/edge/mediapipe/solutions/vision/face_landmarker}} \cite{kartynnik2019realtimefacialsurfacegeometry}, which provides 468 3D facial landmarks per frame, as shown in Fig.~\ref{fig:facemesh} (a). We manually select a subset of 109 key landmark points (omitting redundant or less informative points) to reduce computational complexity while covering the main facial regions (eyes, brows, nose, mouth, jawline, \etc), as shown in Fig.~\ref{fig:facemesh} (b). 

We then normalize the facial landmarks using a frame-based approach. In particular, we take the 3D position of the nose tip landmark shown in red in Fig.~\ref{fig:facemesh} (c), and subtract it from all the other landmark positions in that frame to achieve translation invariance. Then we scale the landmark positions with the intercanthal distance, \ie, the distance between the two inner corners of the eyes, shown in blue in Fig.~\ref{fig:facemesh} (c), resulting in scale invariance.

\subsection{Graph Neural Networks}

A Graph Neural Network (GNN) is a neural architecture designed to process data represented as a graph \(G=(V,E)\), where $V$ is the set of nodes and $E$ the set of edges \cite{miguel25gnn}. Each node $v\in V$ is associated with an initial feature vector $x_v\in\mathbb{R}^d$ (\ie, the 3D coordinate of a landmark, with $d=3$) and, through a series of messages passing layers $L$, the GNN updates the features of each node $h_v^{(l)}$ aggregating information from its neighbors, \ie, nodes connected to it by an edge.

% At layer $k$ of the GNN, the feature of node $i$ is updated as follows (where  $k=1,\dots,L$ denotes the layer index):

% \begin{equation}
% h_i^{(k)}
% = \sigma\Bigl(
% W_{\mathrm{self}}^{(k)}\,h_i^{(k-1)}
% + W_{\mathrm{neigh}}^{(k)} \sum_{j \in \mathcal{N}(i)} h_j^{(k-1)}
% + b^{(k)}
% \Bigr)
% \end{equation}

% where $W_{\mathrm{self}}^{(k)}, W_{\mathrm{neigh}}^{(k)} \in\mathbb{R}^{d^{(k)}\times d^{(k-1)}}$ and $b^{(k)}\in\mathbb{R}^{d^{(k)}}$ are trainable parameters, $\mathcal{N}(i)$ is the set of neighbors of node $i$ and $\sigma(\cdot)$ is an element-wise nonlinearity (\eg ReLU).

By stacking $L$ layers, the representation of each node $h_v^{(L)}$ incorporates information from nodes up to $L$ hops away, captures both the local and global graph structure. 
The final node embeddings can then be pooled (\ie, averaged or summed) to produce a graph‐level embedding, $h_G$.
% \begin{equation}
% h_G = \mathrm{POOL}\bigl(\{\,h_i^{(L)} : i\in V\}\bigr).    
% \end{equation}

GNNs are particularly well suited for facial motion modeling because they explicitly encode the mesh-like geometry of the face through a graph structure, enabling parameter sharing across spatially connected landmarks. This results in both spatial awareness (by preserving local neighborhood relationships among facial regions) and parameter efficiency, given the small and fixed graph size typical of landmark-based representations. In contrast to approaches that rely on high-dimensional pairwise distance tensors (\eg, all-pairs landmark distances as in \cite{prashnani2024avatar}), a graph representation avoids quadratic growth in input size and facilitates faster training and inference.

In particular, we use Graph Convolutional Networks (GCNs), whose message-passing mechanism aggregates information from a node’s local neighbors at each layer. This convolution-like propagation naturally models coordinated facial movements, such as simultaneous eyebrow raises or lip articulations, by allowing information to flow across connected regions. Stacking multiple GCN layers increases the receptive field over the mesh, capturing both local and global facial dynamics.

%These unique properties make GCNs an effective choice for learning per-frame representations that are sensitive to the spatio-structural patterns characteristic of an individual’s facial motion.

\subsection{Proposed System}

Our biometric system follows a spatio-temporal architecture designed to learn identity‐specific facial motion patterns from avatar videos. As illustrated in Fig.~\ref{fig:model}, the core idea is to represent each video as a sequence of facial landmark graphs, process them with a GCN to encode spatial structure, and then aggregate them via a temporal attention-based pooling mechanism. 

% The entire system is trained with a triplet loss to encourage discriminative embeddings that separate genuine and impostor motion signatures.

Each avatar video is divided into non-overlapping fixed-length clips of $T=50$ frames. This choice is motivated by previous work \cite{prashnani2024avatar}, which found that sequences approaching $\approx$30 frames begin to produce a strong verification performance with landmark-based input.

Given an avatar clip of length $T$ frames, we first extract $V=109$ 3D facial landmarks per frame $t$ using MediaPipe \cite{kartynnik2019realtimefacialsurfacegeometry}. For each frame $t$, the landmarks define the nodes $V$ of a graph whose edges $E$ are constructed by Delaunay triangulation \cite{delaunay1934sphere}. The result is a sequence $\{G^{(t)}\}_{t=1}^T$ of graphs that encode both local and global facial structure. 

Each graph $G^{(t)}$ in the sequence is then processed by a 3-layer GCN ($L=3$), in which the first layer transforms the initial feature vector $x_v$ into an embedding $h_v^{(1)}\in\mathbb{R}^{d^{(1)}}$, the second layer transforms $h_v^{(1)}$ into $h_v^{(2)}\in\mathbb{R}^{d^{(2)}}$, and the third layer transforms $h_v^{(2)}$ into $h_v^{(3)}\in\mathbb{R}^{d^{(3)}}$, being $d^{(1)} = d^{(2)} = 64$, and $d^{(3)}=256$. Within each layer, node features are updated by aggregating information from their immediate neighbors, followed by non-linear ReLU activation and dropout ($p=0.3$). Applying the same GCN across all $T$ frames results in a sequence of spatial embeddings $\{h_G^{(t)}\}_{t=1}^T$ that encode the geometry of facial expressions.

To convert this sequence of frame embeddings into a single discriminative descriptor of the entire clip, $e\in \mathbb{R}^{d^{(3)}}$, we consider a temporal attention mechanism. This module learns to assign weights to each embedding at the frame level based on its relevance to identity verification, resulting in higher attention weights in frames with distinctive facial motion patterns and smaller attention weights in less distinctive intervals (\eg, static segments). Attention scores are applied to frame embeddings to compute a weighted sum, resulting in the final clip-level embedding.

\section{Experimental Protocol}\label{sec:experimental_protocol}

For reproducibility reasons, we adopt the same train-validation-test split on the CREMA-D and RAVDESS datasets as in \cite{prashnani2024avatar}. In this protocol, all identities are split disjointly between sets to ensure that no identity is seen in the test set (no targets or drivers) during training or validation. The same identity can only be used as a driver or target within the same split. We refer to these data splits as $\mathcal{D}_{\text{train}}$, $\mathcal{D}_{\text{val}}$, and $\mathcal{D}_{\text{test}}$. Next, we describe the key details for the training and final evaluation protocols.

\subsection{Biometric System: Training}

Our biometric system is trained on $\mathcal{D}_{\text{train}}$, with hyperparameters tuned on $\mathcal{D}_{\text{val}}$. The final model checkpoint is selected on the basis of the lowest validation loss. Training is carried out using triplet loss \cite{morales22circle}. Each triplet $(a, p, n)$ consists of: \textit{i)} an anchor, $a$, which can be any avatar video (\textit{genuine avatar} or \textit{impostor avatar}), \textit{ii)} a positive sample, $p$, which is any avatar video driven by the \emph{same} person as $a$ (regardless of target identity), and \textit{iii)} a negative sample, $n$, which is any avatar video driven by a \emph{different} person than $a$ (again, regardless of target identity). We do not mine hard or semi-hard triplets, but only random triplets. 

The model is implemented in PyTorch, using PyTorch Geometric for the GCN layers. We train on a single NVIDIA RTX 4090 GPU, with typical training times of about 2 hours for 200 epochs, using a batch size of 1,024 and an Adam optimizer with a learning rate of $1e^{-4}$.
 
% This formulation encourages clips driven by the same individual to lie closer in the embedding space than those driven by different individuals.

\subsection{Biometric System: Evaluation}

For the final evaluation of the biometric system, we create verification pairs from $\mathcal{D}_{\text{test}}$ by generating all possible \textit{genuine} and \textit{impostor comparisons} for all identities in $\mathcal{D}_{\text{test}}$, as can be seen in Fig.~\ref{fig:avatares-pares} (right). 

For a given identity $i$ in $\mathcal{D}_{\text{test}}$, we use a neutral high quality image (target image), from which we generate a \textit{ original avatar}. We use all original videos from identity $i$ to generate $N$ \textit{genuine avatar} video clips, which simulate the real usage of avatar $i$. From all those \textit{genuine avatar} clips, we take one as reference clip, and we build all the $N-1$ \textit{genuine comparisons} with the remaining \textit{genuine avatar} clips. We gather all other identities in $\mathcal{D}_{\text{test}}$, and we generate $M$ \textit{impostor avatar} clips using the impostor videos and the target avatar. We then use the same reference \textit{genuine avatar} clip to compare with the $M$ \textit{impostor avatar} clips, obtaining $M$ \textit{impostor comparisons}.

We repeat this process using all identities in $\mathcal{D}_{\text{test}}$, and using each of their \textit{genuine avatar} clips as a reference clip for the comparisons. We use the scores obtained by our biometric system from all the \textit{genuine comparisons} and all the \textit{impostor comparisons} to obtain the performance results.

This protocol explicitly tests if the biometric system can distinguish whether two avatar videos share the same underlying driver despite identical visual appearance. In addition, as $\mathcal{D}_{\text{test}}$ contains identities that are unseen during training or validation, the reported results reflect the generalization capability of the biometric system on new identities.

Similar to \cite{prashnani2024avatar}, we use Area Under the Curve (AUC) as the primary metric for verification performance.

\begin{figure*}[t]
  \centering
  \begin{minipage}[b]{0.32\linewidth}
    \centering
    \includegraphics[width=\linewidth]{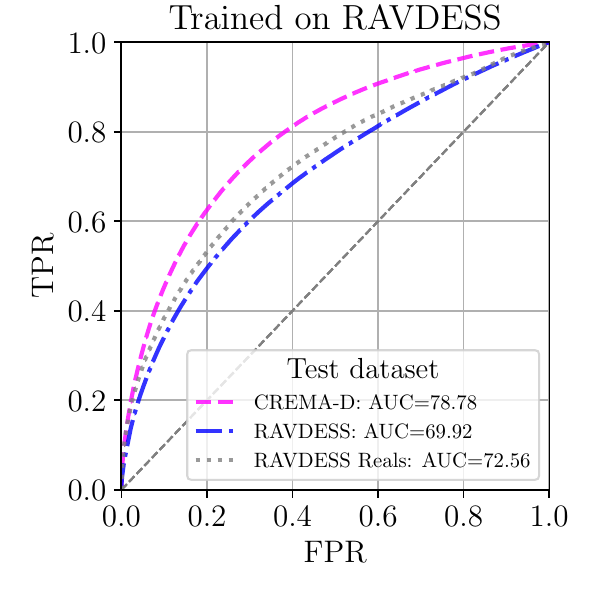}
  \end{minipage}
  \begin{minipage}[b]{0.32\linewidth}
    \centering
    \includegraphics[width=\linewidth]{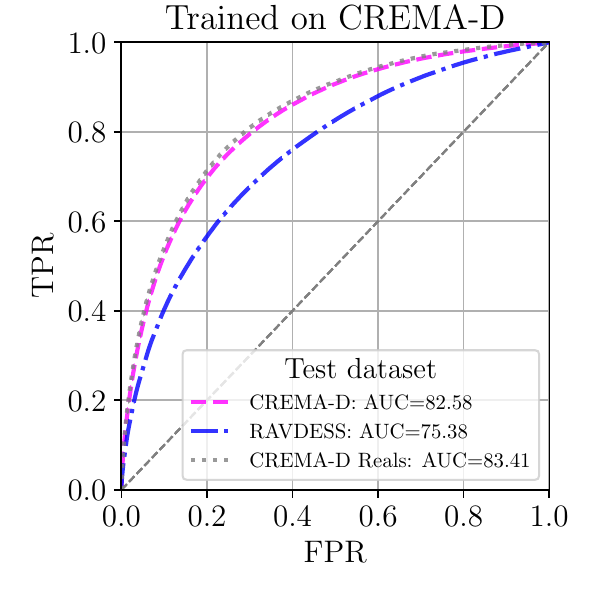}
  \end{minipage}
  \begin{minipage}[b]{0.32\linewidth}
    \centering
    \includegraphics[width=\linewidth]{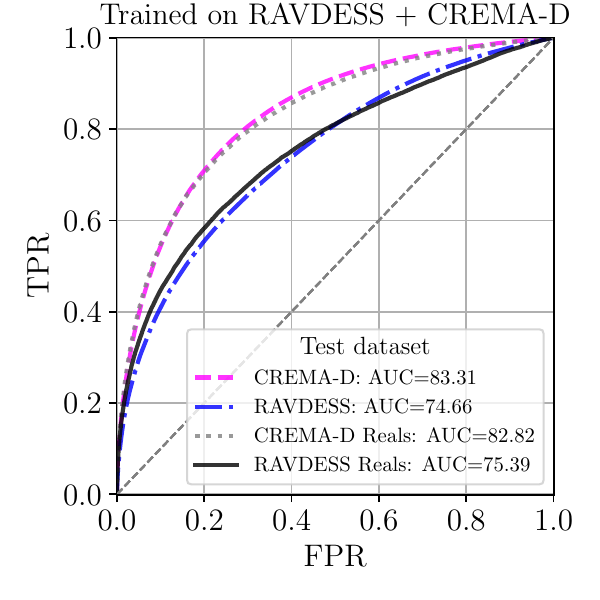}
  \end{minipage}
  \caption{\textbf{Experimental results using our proposed biometric system.} \textit{(Left)} ROC curves obtained when we train our model with avatar clips based on RAVDESS data. \textit{(Center)} ROC curves when we train our model with avatar clips based on CREMA-D data. \textit{(Right)} ROC curves when we train our model with avatar clips based on RAVDESS and CREMA-D data combined.}
  \label{fig:rocs}
\end{figure*}

\section{Experimental Results}\label{sec:experimental_result}

This section evaluates our proposed biometric verification system in multiple settings. All experiments use the same identity-disjoint train-validation-test splits as described in Sec.~\ref{sec:experimental_protocol}, ensuring that identities included in the final evaluation data split $\mathcal{D}_{\text{test}}$ are entirely unseen.

\subsection{Results on Real Data (w/o Avatars)}

First, to validate our proposed biometric verification system, we evaluate the performance of the system on the original test videos (\ie, the real videos without avatar generation). For this experiment, we train our system using avatar videos, and then evaluate on the original videos (without avatars) from different datasets. Because these real videos perfectly preserve each user's natural facial motion, this setup helps us estimate the potential upper-bound performance achievable if the avatar generation process perfectly preserved the original facial motion, \ie, how well our system can discriminate facial motion signatures when no avatar generation artifacts are present.

As can be seen in Fig.~\ref{fig:rocs} (curves labeled ``RAVDESS Reals" and ``CREMA-D Reals"), our biometric system achieves AUC values up to 83\%, demonstrating that the proposed approach is effective for biometric verification using only landmark-based motion cues. 

%The results obtained for RAVDESS real test videos (Fig.~\ref{fig:rocs}, left) improve 3\% when compared to their corresponding RAVDESS avatar videos, while for CREMA-D real test videos (Fig.~\ref{fig:rocs}, center), the improvement is around 1\% when compared to the corresponding CREMA-D avatar videos.

% The RAVDESS set yields slightly lower AUC (75.39\%), likely due to less expressive variation or simpler utterances, which limit the intra-user biometric variability.

% This experiment confirms that our proposed model can learn facial motion patterns, validating the potential of behavioral biometrics in video, even before considering the added challenge of avatars.

\subsection{Intra-Dataset Analysis}

Next, we analyze the virtual meeting scenario using avatars, following the experimental protocol described in Sec.~\ref{sec:experimental_protocol}, focusing on the intra-dataset analysis. That is, we train and test the biometric system on avatar clips derived from the same dataset (CREMA-D or RAVDESS). These intra-dataset experiments simulate a scenario where enrollment and verification stages use avatars generated under similar conditions and from similar populations.

For the RAVDESS dataset, Fig.~\ref{fig:rocs} (left), our system achieves 69.92\% AUC when trained on the same dataset. In the case of CREMA-D dataset, Fig.~\ref{fig:rocs} (center), the AUC is 82.58\%. This higher performance on CREMA-D dataset might be due to: \textit{i}) CREMA-D dataset has a larger number of identities and clips than RAVDESS, which improves training diversity by adding variability to the facial landmark layouts, so the model does not focus in ``absolute" landmark positions; and \textit{ii}) CREMA-D has richer and more varied emotional expressions and speech content, which yield stronger behavioral biometric information.

% These results show that even when appearance is identical (stolen avatar), facial motion patterns enable effective verification.

\subsection{Inter-Dataset Analysis}

We evaluate our biometric system across datasets to assess generalization. We train on one dataset (\eg, RAVDESS $\mathcal{D}_{\text{train}}$) and test on the other (\eg, CREMA-D $\mathcal{D}_{\text{test}}$), evaluating whether the system learns generalizable facial motion representations, or overfits to recording conditions, actor pools, or expression styles of a specific dataset. We follow the same protocol as in Sec.~\ref{sec:experimental_protocol}.

In the case of training with CREMA-D $\mathcal{D}_{\text{train}}$ and evaluating on RAVDESS $\mathcal{D}_{\text{test}}$, Fig.~\ref{fig:rocs} (center), our biometric system achieves 75.38\% AUC, which is lower than its corresponding intra-dataset result for CREMA-D (82.58\% AUC), as expected, while in the case of training with RAVDESS $\mathcal{D}_{\text{train}}$ and evaluating on CREMA-D  $\mathcal{D}_{\text{test}}$, Fig.~\ref{fig:rocs} (left), we obtain 78.78\% AUC, which is higher than its corresponding intra-dataset result for RAVDESS (69.92\% AUC). This counterintuitive result might be explained by the richer expressive variation in the CREMA-D test set. While RAVDESS training enforces more general motion representations due to its limited variability, the CREMA-D test data offers more diverse and distinctive facial motions, making impostor and genuine comparisons more separable in embedding space. Thus, domain shift here actually acts as a form of regularization, and the richer test-time variation improves verification performance.

\subsection{Combined Datasets}

In this experiment, we evaluate whether increasing training data diversity improves verification performance. Specifically, we train our biometric system on the avatar datasets generated from combining both CREMA-D and RAVDESS identities. We hypothesize that by exposing the model to a broader variety of identities, expressions, emotions, and speaking styles, it would help it learn more generalizable facial motion-based representations.

For evaluation, we use the standard $\mathcal{D}_{\text{test}}$ splits of each dataset separately, ensuring no identity overlap with the training data. As can be seen in Fig.~\ref{fig:rocs} (right), we can observe an improvement in AUC compared to their respective intra-dataset results, for both CREMA-D (from 82.58\% to 83.31\% AUC) and RAVDESS (from 69.92\% to 74.66\% AUC) datasets. If evaluated in their corresponding real test datasets, Fig.~\ref{fig:rocs} (right, labeled ``RAVDESS Reals" and ``CREMA-D Reals"), the AUC obtained is very similar, which supports the hypothesis that our system works similarly for avatar videos or real videos.

These gains demonstrate that combining datasets introduces useful variability that enhances the model’s ability to capture consistent, identity-specific facial motion patterns.

%In real-world scenarios, user populations and usage conditions can be highly diverse. Training with richer, more varied data helps the system avoid overfitting to the peculiarities of a single dataset and instead focus on truly generalizable behavioral biometrics.

\begin{figure*}[t]
    \centering
    \includegraphics[width=0.91\linewidth]{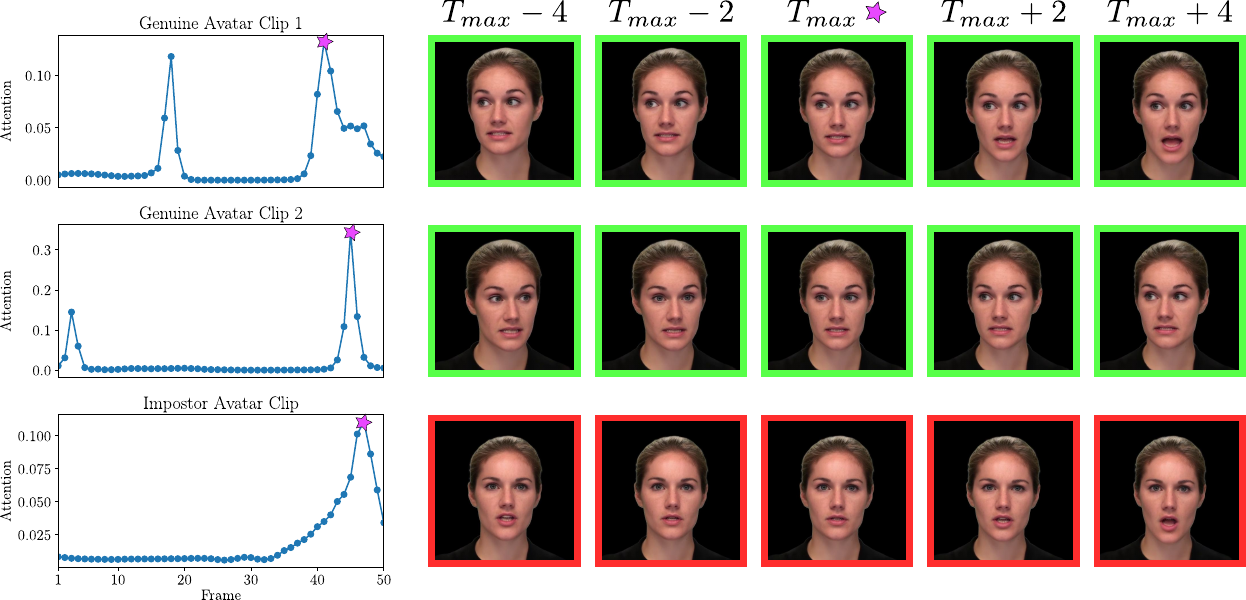}
    \caption{\textbf{Example of attention weights obtained for two \textit{genuine avatar} videos (first and second row) and one \textit{impostor avatar} video (bottom row) for an identity from the RAVDESS dataset:}  The attention weights peak at frame $T_{max}$ \includegraphics[height=0.7em]{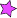}, around frames with characteristic facial gestures from the underlying identity. In this case, the woman shows very similar facial expressions in the frames with highest attention values in both genuine videos. In the third row, the \textit{impostor avatar} is making a facial gesture around frame $T_{max}$ that is probably distinctive from the real underlying impostor identity, not the target identity.} %Attention weights obtained using a model trained on CREMA-D.
    \label{fig:attention-curves}
\end{figure*}

\subsection{Explainability}

To better understand what our biometric system learns, we analyze the temporal attention weights produced by the pooling mechanism. An example of the attention curves over time is shown in Fig.~\ref{fig:attention-curves}, in which we include the attention weight curves and some video frames around the attention peaks for two \textit{genuine avatar} clips and an \textit{impostor avatar} clip for the same target appearance.

Visual inspection of attention curves across multiple clips reveals that the weights tend to peak around frames close to the apex of identity-characteristic expressions, such as wide mouth movements, eyebrow raises, or other distinctive facial gestures or behavioral patterns. These peaks indicate that the system assigns higher importance to frames with rich, discriminative facial motion patterns, while down-weighting more neutral or static frames.

These observations suggest that the model relies on behavioral biometrics rather than the static structure of facial landmarks. Even though the appearance is identical in all avatar videos, the system learns to pick out subtle dynamic patterns unique to each individual.

This behavior confirms our design hypothesis that temporal attention can automatically identify and emphasize the most informative facial motion patterns without requiring predefined expression labels, facial gesture or action units detection, or any alternative method for facial gesture detection. It also provides intuitive explainability, helping users understand which parts are driving verification.

\section{Discussion}\label{sec:discussion}

Our experiments show that facial motion-based biometric verification is feasible even in challenging avatar-mediated scenarios. Across multiple settings and datasets, our proposed biometric system has achieved AUC values that surpass 80\% in the best cases. It is worth noting that our system exclusively focuses on landmarks alone, without using any facial appearance cues or conventional DeepFake detection features. This is a key design choice: in real avatar-mediated communication, all users share the same high-quality rendering pipeline, and appearance artifacts will not distinguish impostors from genuine users. 
%\cite{deandres2025second,melzi2024frcsyn, shahreza2024sdfr} 

Our dataset generation (using \textit{ avatars of impostors} that perfectly match the appearance of the victim) forces the verification model to focus exclusively on behavioral facial motion rather than static facial structure. If we had simply used the original real videos, the model could trivially rely on differences in facial geometry to verify identities. Such structural cues would be useless in a real attack scenario, where a stolen avatar would exactly replicate the victim’s face. By constructing \textit{impostor avatars}, we ensure that the model is trained on the realistic and challenging problem of detecting who is driving the avatar's movements.

Compared to the only other published system for this task \cite{prashnani2024avatar}, which reports up to 87\% AUC using a proprietary model trained on three datasets, our approach remains competitive despite using fewer data and GPU resources. Our biometric system is also lighter with reduced computational complexity, and provides explainability through the temporal attention weights revealing which frames and expressions the model considers most informative. This positive aspect is crucial, as described in the literature \cite{deandres2024pixels,deandres2024good}. 

One limitation of our biometric system is its exclusive reliance on landmark-based representations: while interpretable and efficient, they depend on the quality of the landmark estimator. We have used MediaPipe \cite{kartynnik2019realtimefacialsurfacegeometry}, a widely adopted and robust model that has shown excellent tracking results in comparative studies \cite{luis25run}, but inaccuracies can occur, especially under challenging poses or expressions, potentially reducing verification accuracy.

Another limitation is inherent in the avatar generation process itself. Although GAGAvatar represents the state-of-the-art in talking head one-shot synthesis \cite{chu2024generalizable}, it does not always faithfully reproduce subtle or extreme facial expressions from the driver. Improving avatar generation fidelity, particularly in capturing fine-grained facial dynamics, could directly benefit biometric performance.

% Additionally, while our training strategy focuses solely on driver motion, our experiments also reveal some variation in performance across datasets. This suggests that diversity in expressions, speaker demographics, and recording conditions remains important for generalization.

\section{Conclusions}\label{sec:conclusions}

In this work, we have introduced the challenge of biometric verification in photorealistic talking head avatar videos, where impostors can perfectly impersonate the appearance of a victim. We have proposed a lightweight, explainable system based solely on facial landmarks' motion, using a spatio-temporal GCN with temporal attention-based pooling to learn discriminative behavioral signatures.

In addition to the system code, we have also released a public standard benchmark of \textit{genuine} and \textit{impostor avatar} videos for this scenario. Our experiments have preliminary shown that facial motion patterns are effective biometric cues, demonstrating the potential of behavioral biometrics as a defense against avatar-based impersonation. Future work will be oriented to improve the performance through novel deep learning architectures \cite{delgado2023exploring, goswami2024moment, Marija_Ivanovska_FG2025} and loss functions \cite{boutros2022elasticface,morales22circle,gonzalez2025type2branch} and the evaluation of the biometric system under more challenging scenarios and avatar generation models \cite{chu2024generalizable, li2025rgbavatar, liu2025avatarartist, xu2024gaussian}. Future work will also integrate other information beyond facial cues in multimodal approaches \cite{pena23mm} with speech and other interaction signals \cite{acien20bots,acien22bots}.

\section*{Acknowledgements}
\label{sec:acks}

This study has received funding from INTER-ACTION (PID2021-126521OBI00 MICINN/FEDER), HumanCAIC (TED2021-131787B-I00 MICINN), Cátedra ENIA UAM-VERIDAS en IA Responsable (NextGenerationEU PRTR TSI-100927-2023-2), and PowerAI+ (SI4/PJI/2024-00062 Comunidad de Madrid and UAM).

% \section{DO NOT FORGET!!!}

% Page limit should be  8 pages of main contents (including figures and tables) + extra pages for references.

% Add numbers to all equations and sections and figures, regardless of being explicitly mentioned in the paper.

% DO NOT use "our", or "my" when citing preivous work

% DO NOT add acknowledgements in the review version!

% If there are 3 or more authors, use "et al.".    Use \eg  and \etal 

% \section{Only for Final copy}

% You must include your signed IEEE copyright release form when you submit
% your finished paper. We MUST have this form before your paper can be
% published in the proceedings.

{\small
\bibliographystyle{ieee}
\bibliography{egbib}
}

\end{document}